\documentclass[10pt,twocolumn,letterpaper]{article}
\usepackage[
  letterpaper,
  textwidth=6.875in,
  textheight=8.875in,
  top=1in
]{geometry}
\usepackage{amsmath,amssymb,booktabs,multirow,graphicx,microtype}
\usepackage{times}
\usepackage{array,tabularx,makecell,enumitem}
\usepackage{float}
\usepackage{cuted}
\usepackage[hypcap=false]{caption}
\usepackage{xspace}
\usepackage[numbers,sort&compress]{natbib}
\usepackage{xcolor}
\usepackage[colorlinks=true,linkcolor=black,citecolor=blue,urlcolor=blue]{hyperref}

\setlist{nosep,leftmargin=*}

\newcommand{\method}{NIF\xspace}
\newcommand{\sg}{\operatorname{sg}}
\newcommand{\clip}{\operatorname{clip}}
\newcommand{\diag}{\operatorname{diag}}
\newcommand{\R}{\mathbb{R}}
\DeclareMathOperator*{\argmin}{arg\,min}

\title{NIF: Normal-Intersection Fields for\\
Geometrically Consistent Oriented Box Refinement}

\author{Shaokai Zheng\thanks{Corresponding author.}\\
Shanghai Zhongchao Automation Engineering Co., Ltd.\\
Shanghai, China\\
\texttt{zsk@zc-sh.net}}
\date{}

\begin{document}
\maketitle
\begin{abstract}
Accurate oriented detection requires boxes to follow object boundaries, yet
small errors in the coupled center, size, and angle estimates can displace
long boundaries even when an object is correctly classified. We propose
\method, a plug-in refinement branch that represents box geometry with
\textbf{\textit{normal-intersection fields}}. Given a coarse oriented box, it samples
local features at 20 boundary anchors and predicts signed distances along
their outward normals. These observations have an exact affine representation
in five rectangle parameters; a differentiable projection uses error-aware
reliability weights to recover a geometrically consistent box.
Tangential pooling preserves the normal-position profile, while the shared
projection adds 0.132M parameters. On DOTA-v1.0,
NIF obtains 77.98\%, 79.03\%, and 79.32\% mAP with Oriented
R-CNN, Strip R-CNN, and LSKNet-S, respectively; LSKNet-S
gives the strongest single-scale result among the compared methods. NIF
also reaches 68.19\% mAP on DIOR-R
and improves matched FAIR1M-v1.0 and DIOR-R baselines by 0.30\% and 0.71\%,
respectively. On HRSC2016, it raises mean IoU from 0.867 to 0.888 while
reducing field error by 24.5\%. Finally, with four different detector
frameworks frozen, NIF improves relative AP$_{75}$ by 1.55--4.16\% and
COCO-style mAP by 0.37--3.76\%, without detector retraining.
Upon acceptance of this paper, we will publicly release the code at
\url{https://github.com/zsk466991125-collab/Normal-Intersection-Fields}.
\end{abstract}

\section{Introduction}
\label{sec:intro}

A ship can be recognized correctly while its predicted rectangle still cuts
across the bow or extends into the water. For a long object, a small angular
error moves the end of a boundary substantially; for a narrow object, a small
normal displacement can remove a considerable fraction of the overlap.
Such errors matter in aerial imagery, where ships, vehicles, bridges, and
aircraft occur at arbitrary orientations and widely varying aspect
ratios~\cite{dota,dota_original,hrsc,dior,fair1m,small_object_survey,rsod_survey}.
Benchmarks such as DOTA, DIOR-R, and FAIR1M reflect dense scenes, small
instances, and fine-grained categories~\cite{dota_original,dior,aopg,fair1m}.
Improving recognition alone does not ensure that the
four sides form an accurate enclosure. In two-stage detectors built on
region-based pipelines~\cite{fasterrcnn}, this motivates a question: \emph{can
an oriented detector correct its final box by observing the boundaries that
the box is intended to represent?}

Recent detectors strengthen localization through oriented RoI alignment
~\cite{roi_transformer,orcnn}, rotation-aware features~\cite{redet,arc,gra},
context and anisotropic kernels~\cite{lsknet,pkinet,pkinetv2,striprcnn}, or
frequency-domain alignment~\cite{faa}. These representations improve the
features, but the final box is still usually decoded from a compact region
descriptor; the link between boundary evidence and joint center, scale, and
angle correction remains implicit.

An additional regression stage improves localization but asks another mapping
to infer the same coupled parameters. Boundary-aware localization and
progressive box refinement have been explored in prior work~\cite{sabl,pbrnet};
we therefore do not claim the first use of boundary cues for localization.
NIF instead samples distributed signed displacements along the four normals
of an existing coarse box and fits them under an explicit rectangle
compatibility constraint. The observations remain defined beyond finite
sides and under equivalent width--height conventions, while the joint fit
prevents independent predictions from disagreeing on a rectangle.

\begin{figure*}[t]
  \centering
  \includegraphics[width=\textwidth]{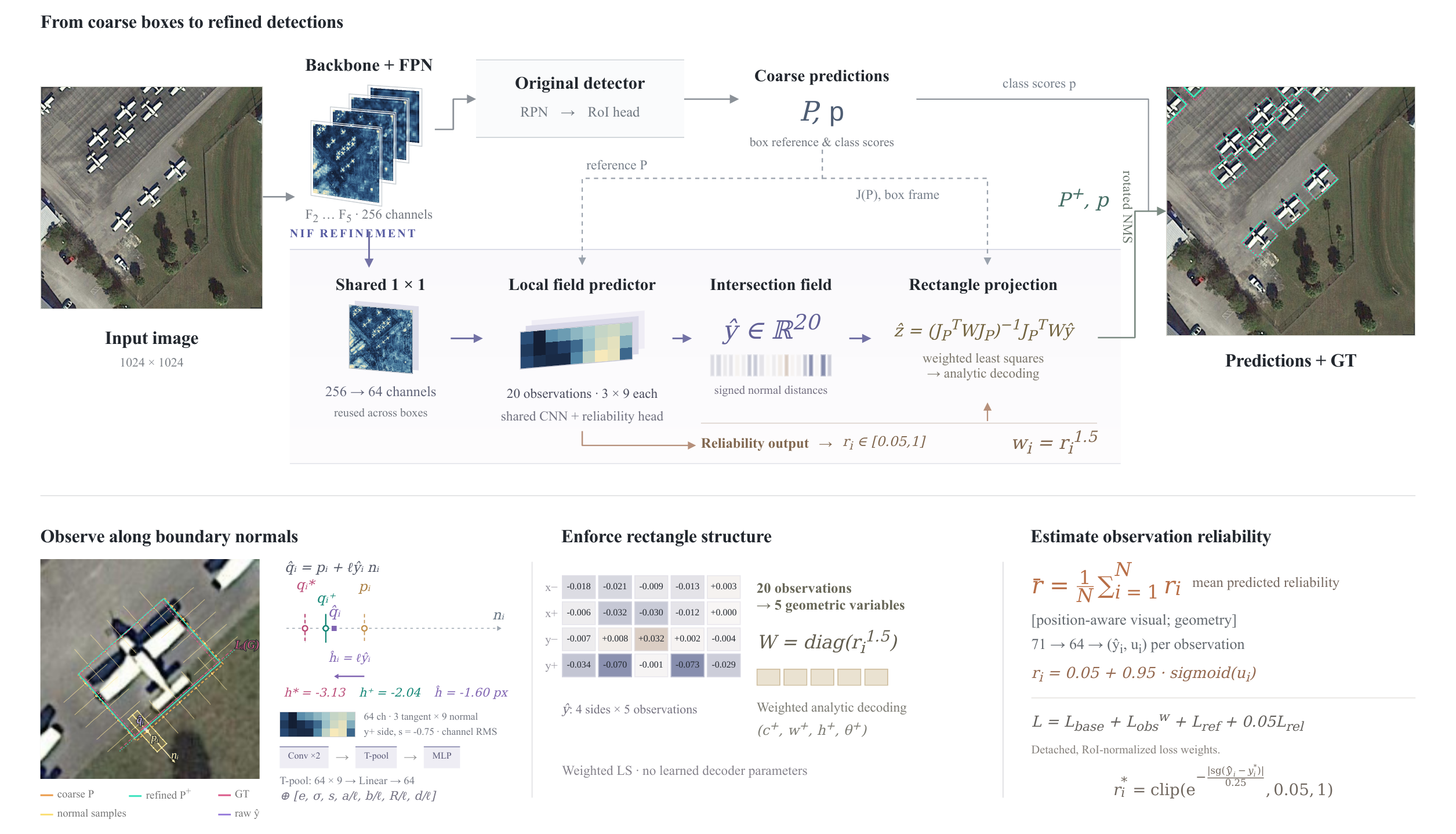}
  \caption{\textbf{Overview of \method.} A carrier predicts coarse boxes
  $P$ and scores. Shared projection and twenty $3\!\times\!9$ normal strips
  predict observations and reliabilities; weighted fitting and analytic
  decoding refine boxes before rotated NMS. The example contrasts raw and
  compatible intersections.}
  \label{fig:framework}
\end{figure*}

We address these issues with \emph{normal-intersection fields} (NIF). From a
coarse rectangle, five anchors per side predict signed displacements along
outward normals to the target supporting lines. After axis alignment, the
20-dimensional field is exactly affine in five rectangle variables: opposite
sides encode translation and scale, and their slopes encode rotation.
Compatibility can therefore be imposed by an explicit fit rather than an
unconstrained output head. Figure~\ref{fig:framework} summarizes the
pipeline.

The local predictor pools only along the tangent, retaining the ordered
normal profile that full-strip pooling would discard. Each descriptor predicts
a displacement and bounded reliability; this reliability is an error-aware
fitting weight, while detached, per-RoI normalized weights are used for
supervision. A shared projection is reused across
boxes, yielding a plug-in with 0.132M additional parameters that preserves the carrier's
backbone, proposal generator, and RoI heads.

We evaluate NIF on DOTA-v1.0, FAIR1M-v1.0, DIOR-R, and HRSC2016. On DOTA,
it reaches 77.98\%, 78.18\%, 79.03\%, and 79.32\% mAP with Oriented
R-CNN, PKINet-v2-S, Strip R-CNN, and LSKNet-S, respectively; LSKNet-S gives
the strongest single-scale comparison. NIF also reaches 68.19\% on DIOR-R.
Gains persist on FAIR1M and HRSC at strict overlap.
With four frozen detectors, training only NIF improves AP$_{75}$ and
COCO-style mAP; field-error and paired-IoU analyses link these gains to
geometric correction.

Our contributions are threefold:
\begin{itemize}
  \item We formulate refinement as a normal-intersection field with an exact
  five-variable affine representation and compatible rectangle fit.
  \item We develop a normal-position-aware predictor with an error-aware
  fitting-weight branch as a post-regression module with 0.132M additional parameters.
  \item We establish SOTA comparisons, cross-framework transfer, and
  geometric evidence linking field compatibility to detection gains.
\end{itemize}

\section{Related Work}
\label{sec:related}

\paragraph{Oriented representations and regression.}
Rotated proposals and aligned RoIs extend region-based detectors to arbitrary
orientations~\cite{fasterrcnn,rrpn,r2cnn,roi_transformer,orcnn}, improving
proposal geometry or pooling while retaining center--size--angle regression.
Boundary vectors, midlines, gliding vertices, adaptive points, and periodic
angle labels reduce encoding ambiguity~\cite{bbavector,middle_lines,gliding,reppoints,csl,dle};
differentiable overlap and Gaussian losses offer alternative objectives
~\cite{piou,gwd,kld,kfiou,grep,giou}. Side-aware localization and pyramidal
refinement use boundary cues, while projection-based parameterization
addresses angular uncertainty~\cite{sabl,pbrnet,projbb}. NIF instead fits
distributed signed normal intersections sampled from a coarse box to a
compatible rectangle with five degrees of freedom. Weak-label methods infer orientation from weaker
annotations~\cite{h2rbox,h2rboxv2,pointobb,point2rbox}, whereas NIF assumes
oriented-box supervision; denoising methods target clutter
~\cite{scrdet,scrdetpp,maskobb}, while NIF enforces side consistency after
detection. Unlike independent coordinate predictions, NIF predicts local
boundary evidence and explicitly projects it onto a shared rectangle geometry.

\paragraph{Alignment, refinement, and features.}
RoI Transformer, S$^2$A-Net, R$^3$Det, Cascade R-CNN, DRN, AOPG, and dynamic
anchors improve alignment, proposals, or staged localization
~\cite{roi_transformer,s2anet,r3det,cascade_rcnn,drn,aopg,dynamic_anchor};
their refinement stages commonly predict another box update or re-align
features. NIF instead fits distributed side evidence to a compatible box in
closed form, without another unconstrained box update. Oriented R-CNN is an
efficient carrier~\cite{orcnn}; FAA aligns
directional features~\cite{faa}; and RIFD, ReDet, ARC, GRA, ReDiffDet,
LSKNet, PKINet, and Strip R-CNN model rotation or anisotropic context
~\cite{rifd,redet,arc,gra,rediffdet,lsknet,pkinet,pkinetv2,striprcnn}.
Large-kernel designs, spatial transformers, and deformable convolution offer
additional alignment mechanisms~\cite{replknet,unireplknet,convnext,stn,dcn}.
NIF is orthogonal to these carriers: it samples each side in its normal frame
and maps the measurements to one geometric correction. Unlike quality
estimators that rank boxes~\cite{iou_net,gfl,varifocal}, its reliability
weights local observations while retaining class scores. This plug-in branch
can be trained jointly or with its carrier frozen, without changing the
detector's prediction head.

\section{Methods}
\label{sec:method}

Given a detector's final coarse oriented box, NIF predicts signed normal
displacements and reliabilities at 20 side anchors. Reliability acts as an
error-aware fitting weight when the observations are projected onto the
five-dimensional family of valid
rectangles, and an analytic decoder returns the refined box. We first derive
the field geometry and compatible fit, then describe the local feature branch,
its supervision, and its integration with the detector.

\subsection{From Box Parameters to Boundary Measurements}
\label{sec:field}

Let $P=(\mathbf c,2a,2b,\theta)$ be a coarse oriented box, where
$\mathbf c\in\R^2$ is its image-plane center, $a,b>0$ are the half-width
and half-height, and $\theta$ is its width-axis angle in radians. Its unit axes are
$\mathbf u=(\cos\theta,\sin\theta)^\top$ and
$\mathbf v=(-\sin\theta,\cos\theta)^\top$. We use the image-coordinate
rotation convention of the carrier and normalize distances by the
half-diagonal $\ell=\sqrt{a^2+b^2}$. On each side, $m=5$ equally spaced positions
$s_j\in[-0.75,0.75]$ define anchors
\begin{align}
 \mathbf p_{x,\sigma,j}&=\mathbf c+\sigma a\mathbf u+s_j b\mathbf v,\nonumber\\
 \mathbf p_{y,\sigma,j}&=\mathbf c+s_j a\mathbf u+\sigma b\mathbf v,
 \label{eq:anchors}
\end{align}
where $j=1,\ldots,m$ indexes positions along a side, $s_j$ is the
dimensionless tangential position, and $\sigma\in\{-1,+1\}$ selects
opposite sides. Subscripts $x$ and $y$ identify sides normal to
$\mathbf u$ and $\mathbf v$, respectively; $\mathbf p$ is an anchor's
image coordinate. The outward unit normals are
$\mathbf n_{x,\sigma}=\sigma\mathbf u$ and
$\mathbf n_{y,\sigma}=\sigma\mathbf v$.

For a matched ground-truth rectangle $G$, let $i=1,\ldots,4m$ enumerate
all $4m=20$ side--position pairs and $L_i(G)$ denote the corresponding target
supporting line. Its signed normal
intersection satisfies
\begin{equation}
 \mathbf q_i^*=\mathbf p_i+h_i^*\mathbf n_i\in L_i(G),
 \qquad y_i^*=h_i^*/\ell.
 \label{eq:intersection}
\end{equation}
Here, $\mathbf p_i$ and $\mathbf n_i$ are the anchor and its outward normal,
$\mathbf q_i^*$ is the intersection, $h_i^*$ is its signed distance in
pixels, and $y_i^*$ is the normalized target observation. Positive $h_i^*$
indicates an outward correction. A star marks a target; hats and tildes
mark predictions and compatible projections, respectively.
Supporting lines extend finite target edges, so intersections remain
defined beyond their endpoints. All targets follow from oriented box
annotations.

\begin{figure*}[t]
  \centering
  \includegraphics[width=\textwidth]{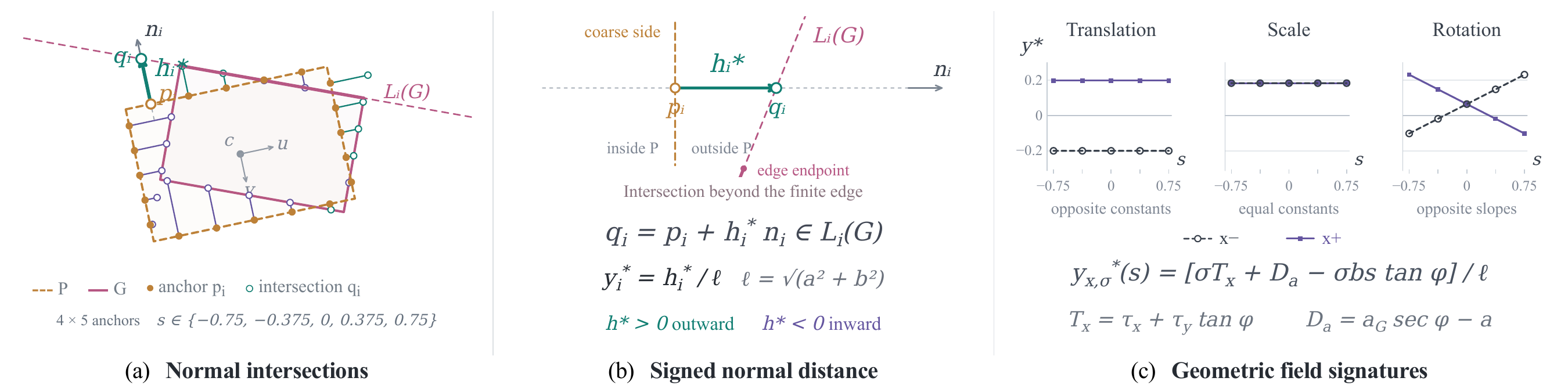}
  \caption{\textbf{Geometry of the normal-intersection field.} (a) Anchors
  and outward normals are attached to the coarse rectangle. (b) The signed
  displacement ends at a corresponding target supporting line, including
  intersections beyond finite edges. (c) Translation, scale, and rotation
  induce different, coordinated constant and slope patterns. These
  patterns motivate an exact five-dimensional compatible field.
  Intersection labels $\mathbf q_i$ denote target points $\mathbf q_i^*$;
  $\mathbf c,\mathbf u,\mathbf v$ define the coarse-box frame.}
  \label{fig:geometry}
\end{figure*}

The target rectangle has equivalent width--height assignments. We choose
the assignment whose relative angle $\phi=\theta_G-\theta$ lies in
$[-\pi/4,\pi/4)$, swapping target half-sizes when needed. This preserves
the rectangle and ensures that $\cos\phi$ is nonzero. In the coarse frame,
let $\tau_x=(\mathbf c_G-\mathbf c)^\top\mathbf u$ and
$\tau_y=(\mathbf c_G-\mathbf c)^\top\mathbf v$ be the center offset,
where $\mathbf c_G$, $a_G,b_G$, and $\theta_G$ are the center, half-sizes,
and orientation of the aligned target. Define
\begin{align}
 t&=\tan\phi,\quad T_x=\tau_x+t\tau_y,\quad T_y=\tau_y-t\tau_x,
 \nonumber\\
 D_a&=a_G\sec\phi-a,\qquad D_b=b_G\sec\phi-b.
 \label{eq:auxiliary}
\end{align}
Here, $t$ encodes relative rotation; $T_x,T_y$ mix translation and rotation,
while $D_a,D_b$ encode half-size changes after angular compensation.
The operator $\sec\phi=1/\cos\phi$. Intersecting the anchor-normal lines
with target supporting lines gives
\begin{align}
 y_{x,\sigma}^*(s)&=(\sigma T_x+D_a-\sigma b s t)/\ell,\nonumber\\
 y_{y,\sigma}^*(s)&=(\sigma T_y+D_b+\sigma a s t)/\ell.
 \label{eq:field}
\end{align}
Here, $s$ is a normalized tangential coordinate, evaluated at $s=s_j$
for the sampled field. This expression is exact on the selected axis
branch, without a small-angle approximation. The constants and slopes in
Fig.~\ref{fig:geometry} explicitly couple the four sides. Learning 20
local observations can therefore be separated from enforcing the five
degrees of freedom of a rectangle.

\subsection{A Five-Variable Compatible Field}
\label{sec:projection}

Let $\kappa=\sqrt{m^{-1}\sum_{j=1}^m s_j^2}$ be the root-mean-square
anchor position. Introduce the normalized variables
\begin{equation}
 \mathbf z=\big(T_x/\ell,T_y/\ell,D_a/\ell,D_b/\ell,\kappa t\big)^\top.
 \label{eq:z}
\end{equation}
The vector $\mathbf z\in\R^5$ contains two translation, two size, and
one rotation variable. Stacking observations in $x-,x+,y-,y+$ order gives
$\mathbf y^*\in\R^{4m}$ and $\mathbf y^*=J(P)\mathbf z$, where
$J(P)\in\R^{4m\times5}$ is the geometry-dependent design matrix with rows
\begin{align}
 J_{x,\sigma,j}&=\big(\sigma,0,1,0,-\sigma b s_j/(\ell\kappa)\big),\nonumber\\
 J_{y,\sigma,j}&=\big(0,\sigma,0,1,\sigma a s_j/(\ell\kappa)\big).
 \label{eq:design}
\end{align}
Each row maps $\mathbf z$ to one observation indexed by side and position.
We write $J=J(P)$ below. Rectangle fields occupy its five-dimensional
column space. Symmetric sampling gives $J^\top J=2mI_5$, where $I_5$ is
the $5\!\times\!5$ identity: odd cross-moments cancel and
$a^2+b^2=\ell^2$ normalizes the rotational column. The design is full
rank for positive box sizes and at least two distinct sample positions.
This also clarifies why scale normalization matters: the five fitted
variables have comparable units before observation weighting.

The network predicts normalized observations
$\hat{\mathbf y}\in\R^{4m}$ and reliabilities $r_i\in(0.05,1)$.
We fit
\begin{align}
 \hat{\mathbf z}
 &=\argmin_{\mathbf z}\sum_{i=1}^{4m} w_i\big((J\mathbf z)_i-\hat y_i\big)^2,
 \nonumber\\
 &= (J^\top WJ)^{-1}J^\top W\hat{\mathbf y},\quad
 W=\diag(w_1,\ldots,w_{4m}).
 \label{eq:wls}
\end{align}
Here, $w_i=r_i^\gamma$ weights observation $i$, $\gamma=1.5$ controls
reliability emphasis, and $\diag$ forms the diagonal matrix
$W\in\R^{4m\times4m}$. The fitted vector is
$\hat{\mathbf z}\in\R^5$. A differentiable linear solve is applied to
the $5\!\times\!5$ normal system. We normalize weights to mean one within
each RoI, which leaves the minimizer unchanged. Positive reliabilities
retain full rank. The result $\tilde{\mathbf y}=J\hat{\mathbf z}$ is the
weighted projection onto a rectangle-compatible supporting-line field.
This fit minimizes the algebraic normal-displacement residual, rather
than the perpendicular distance to a rotated line or a rotated IoU loss.

\paragraph{Analytic rectangle decoding.}
Write $\hat{\mathbf z}=(z_x,z_y,z_a,z_b,z_t)^\top$, where the first two
entries encode translation, the next two encode size, and $z_t$ encodes
rotation. Let $\hat t=z_t/\kappa$ be the fitted angle tangent. The
refined local center offset is
\begin{equation}
 \Delta\mathbf c_{\mathrm{local}}=
 \frac{\ell}{1+\hat t^2}
 \begin{pmatrix}z_x-\hat t z_y\\\hat t z_x+z_y\end{pmatrix},
 \label{eq:center}
\end{equation}
Its components $(\Delta c_x,\Delta c_y)$ are translations in pixels
along $\mathbf u$ and $\mathbf v$. The refined box is obtained by
\begin{align}
 \mathbf c^+&=\mathbf c+\Delta c_x\mathbf u+\Delta c_y\mathbf v,\nonumber\\
 (a^+,b^+)&=(a+\ell z_a,b+\ell z_b)/\sqrt{1+\hat t^2},\nonumber\\
 \theta^+&=\theta+\arctan\hat t.
 \label{eq:decode}
\end{align}
Here, $P^+=(\mathbf c^+,2a^+,2b^+,\theta^+)$ is the refined rectangle;
superscript $+$ denotes its updated center, half-sizes, and angle.
We clamp full sizes to at least $10^{-3}$ pixels and canonicalize to the
carrier's long-edge convention. The fit enforces linear field structure;
clamping ensures positive sizes. A zero field gives the identity.
Exact recovery assumes noise-free targets, consistent side assignments,
and inactive clamping. The supplementary material details the recovery
conditions and reports clamp frequencies.

\subsection{Local Observation Network}
\label{sec:network}

\begin{figure}[t]
  \centering
  \includegraphics[width=\columnwidth]{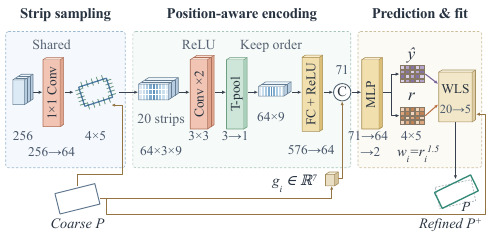}
  \caption{\textbf{Local observation network.} Tensor sizes are per
  observation. T-pool averages along the tangent while preserving nine
  normal positions; C concatenates visual and geometric descriptors.
  The shared predictor outputs 20 field values and reliabilities per box.
  WLS fits five geometric variables for analytic rectangle decoding.}
  \label{fig:network}
\end{figure}

\paragraph{Shared projection and normal strips.}
Figure~\ref{fig:network} details the local branch. A shared
$1\!\times\!1$ convolution reduces 256-channel FPN maps to 64 channels,
reusing each projected map across its assigned boxes. At anchor
$\mathbf p_i$, bilinear sampling forms a $3\!\times\!9$ tangent--normal
grid: nine equally spaced normal offsets span $[-R_i,R_i]$, and three
tangent offsets are $-d,0,d$. Here, $d$ is the assigned FPN level's stride
in image pixels; the tangent is $\mathbf v$ on $x$-sides and $\mathbf u$
on $y$-sides. The normal radius is $R_i=\max(a/2,2d)$ or
$\max(b/2,2d)$, respectively. Sampling follows the carrier's half-pixel
alignment and does not bound the predicted displacement.

\paragraph{Preserving the normal profile.}
The strip encoder in Fig.~\ref{fig:network} applies two
$3\!\times\!3$ Conv--ReLU layers, averages only along the tangent, and
maps the ordered $64\!\times\!9$ profile to a 64-dimensional visual
descriptor. This preserves normal-position cues lost by full spatial
pooling. We concatenate $\mathbf g_i\in\R^7$, containing the side type
($x=0$, $y=1$), side sign $\sigma_i$, position $s_i$, and four quantities
$a,b,R_i,d$ normalized by $\ell$. The shared MLP predicts the field value
$\hat y_i$ and reliability logit $u_i$. We set
$r_i=r_{\min}+(1-r_{\min})\operatorname{sigmoid}(u_i)$, where
$r_{\min}=0.05$ bounds reliability away from zero and sigmoid maps a
logit to $(0,1)$. All sides, anchors, pyramid levels, and classes share
the predictor. Zero-initialized field outputs and initial reliability
0.8 make the initial box update the identity.

\subsection{Supervision and Detector Integration}
\label{sec:training}

NIF is a plug-in branch after coarse-box regression: it reuses pyramid
features and box coordinates while preserving the carrier's backbone, RPN,
RoI heads, and proposal assignment. For each foreground RoI, we detach the
coarse-box coordinates and construct $\mathbf y^*$ from its matched GT.
Original detector losses remain active. NIF adds
\begin{equation}
 \mathcal L=\mathcal L_{\mathrm{base}}
 +\mathcal L_{\mathrm{obs}}+\mathcal L_{\mathrm{ref}}
 +\lambda_{\mathrm{rel}}\mathcal L_{\mathrm{rel}}.
 \label{eq:total_loss}
\end{equation}
Here, $\mathcal L_{\mathrm{base}}$ sums the carrier's original RPN and
RoI classification and regression losses. $\mathcal L_{\mathrm{obs}}$,
$\mathcal L_{\mathrm{ref}}$, and $\mathcal L_{\mathrm{rel}}$ supervise the
field, refined box, and reliability, respectively. Their weights are
$1$, $1$, and $\lambda_{\mathrm{rel}}=0.05$. Define
\begin{align}
 \mathcal L_{\mathrm{obs}}
 &=\mathbb E_P\!\left[\frac1{4m}\sum_{i=1}^{4m}
  \bar w_i(\hat y_i-y_i^*)^2\right],\nonumber\\
 \bar w_i&=\frac{\sg(r_i^\gamma)}
 {\frac1{4m}\sum_{i'=1}^{4m}\sg(r_{i'}^\gamma)},
 \label{eq:observation_loss}\\
 \mathcal L_{\mathrm{ref}}&=\mathbb E_{P,G}[1-\operatorname{rIoU}(P^+,G)].
 \label{eq:refinement_loss}
\end{align}
Here, $\mathbb E_P$ and $\mathbb E_{P,G}$ denote empirical means over
selected RoIs and their matched GT pairs. The scalar $\bar w_i$ is the
per-RoI normalized observation weight; $i'$ enumerates all observations
in its denominator. $\sg$ is stop-gradient: it preserves a value in the
forward pass and blocks its backward derivative. The operator
$\operatorname{rIoU}(P^+,G)$ is intersection area divided by union area
for the two rotated rectangles. Detached weights prevent the observation
loss from directly lowering reliability; per-RoI normalization removes
uniform weight scaling. The fit remains differentiable in both outputs.
We use exact polygon-based rotated IoU with piecewise differentiable
intersection geometry.

Reliability is supervised by the current detached observation error:
\begin{align}
 r_i^*&=\clip\!\left(
 \exp\!\left[-\frac{|\sg(\hat y_i)-y_i^*|}{\tau_r}\right],r_{\min},1\right),
 \label{eq:reliability_target}\\
 \mathcal L_{\mathrm{rel}}
 &=\mathbb E_P\!\left[\frac1{4m}\sum_{i=1}^{4m}
 \operatorname{SmoothL1}_{\beta_r}(r_i-r_i^*)\right].
 \label{eq:reliability_loss}
\end{align}
Here, $r_i^*$ is the error-derived reliability target and $\tau_r=0.25$
sets its decay scale in normalized displacement units.
$\clip(x,L,U)=\min(\max(x,L),U)$ clips input $x$ between lower bound
$L$ and upper bound $U$, here $r_{\min}$ and $1$.
For residual $e=r_i-r_i^*$, $\operatorname{SmoothL1}_{\beta_r}(e)$ equals
$e^2/(2\beta_r)$ when $|e|<\beta_r$ and $|e|-\beta_r/2$ otherwise,
with transition $\beta_r=0.1$. We use up to 128
detector positives per image, excluding GT-injected proposals when other
positives are available.
NIF uses no quality-only negatives or synthetic perturbation supervision.
Detaching reference boxes isolates the sampling geometry, while the
local network and shared backbone/FPN still receive NIF gradients.

At inference, NIF corrects candidates after the carrier's score filtering
and cap, before rotated NMS; class scores remain unchanged. It adds no
proposal stage or iterative loop.

\section{Experiments}
\label{sec:experiments}

\subsection{Experimental Settings}

\paragraph{Datasets and metrics.}
DOTA-v1.0~\cite{dota} contains 2,806 aerial images and 15 categories,
with 1,411/458/937 images in train/val/test. We train on trainval and
report the official test-set mAP at IoU 0.5. Images are cropped into
$1024\!\times\!1024$ patches with a 200-pixel overlap and a single scale
factor of 1.0. Patch predictions are mapped back and merged
for submission.
HRSC2016~\cite{hrsc} uses the standard 617 trainval and 444 test images;
the test pipeline preserves aspect ratio when resizing the longer side to 800.
We report AP$_{50}$, AP$_{75}$, and COCO-style mAP using 11-point VOC 2007
and all-points VOC 2012 interpolation~\cite{voc}.
FAIR1M-v1.0~\cite{fair1m} contains 37 aerial object categories.
We use its official training and test splits and report the official mAP
over 34 fine-grained categories. Following Strip R-CNN~\cite{striprcnn},
images are rescaled by $\{0.5,1.0,1.5\}$ and cropped into
$1024\!\times\!1024$ patches with a 500-pixel overlap for both training
and testing. Predictions from all scales are mapped back and merged
before submission.
DIOR-R~\cite{aopg} adds oriented annotations to 20 categories in DIOR.
We use the native $800\!\times\!800$ images, train on trainval, and
evaluate on test with the native rotated-IoU $0.5$ VOC 2007 eleven-point
mAP protocol.

\paragraph{Implementation.}
All experiments are implemented with the MMRotate toolbox~\cite{mmrotate}.
The experiments use 256-channel FPN features~\cite{fpn}, ImageNet-pretrained
backbones, including ResNet~\cite{resnet}, and the original carrier heads;
NIF is appended after the final coarse box regression. Following
FAA~\cite{faa}, DOTA-v1.0 uses $1024\!\times\!1024$ inputs, while
HRSC2016 and DIOR-R use an 800-pixel target size. Training uses AdamW~\cite{adamw} with an initial
learning rate of $10^{-4}$, weight decay 0.05, and
$(\beta_1,\beta_2)=(0.9,0.999)$. The batch size is two per GPU on four
GPUs without learning-rate scaling. For DOTA-v1.0 and HRSC2016, we train
for 36 epochs, reduce the
learning rate by a factor of ten at epochs 24 and 33, and use a 500-iteration
linear warm-up with an initial ratio of $1/3$. DIOR-R and FAIR1M use 12 epochs
with learning-rate reductions at epochs 8 and 11. All datasets use trainval
for training and the corresponding official test split for evaluation. NIF
hyperparameters follow Sec.~\ref{sec:method}. Complexity and latency in
Tables~\ref{tab:hrsc} and~\ref{tab:dior} use the complete test splits,
FP32, batch size one, and one RTX 2080 Ti; ten warm-up images are discarded.
Synchronized latency covers the end-to-end forward, including rotated
RoIAlign, strip sampling, NIF fitting, decoding, and NMS, while loading,
evaluation, and writing are excluded. GFLOPs count supported MMCV modules;
unsupported operators and tensor-level geometry affect latency but not GFLOPs.
HRSC preserves aspect ratio and DIOR uses native $800\times800$ images.

\subsection{Detection Results}
\label{sec:main_results}

\begin{table*}[t]
\centering
\scriptsize
\setlength{\tabcolsep}{2.35pt}
\begin{tabular*}{\textwidth}{@{\extracolsep{\fill}}lcccccccccccccccc}
\toprule
Method & PL & BD & BR & GTF & SV & LV & SH & TC & BC & ST & SBF & RA & HA & SP & HC & mAP \\
\midrule
R. Faster R-CNN~\cite{fasterrcnn} & 89.40 & 81.81 & 47.28 & 67.44 & 73.96 & 73.12 & 85.03 & \textbf{90.90} & 85.15 & 84.90 & 56.60 & 64.77 & 64.70 & 70.28 & 62.22 & 73.17 \\
R$^3$Det~\cite{r3det} & 89.50 & 81.20 & 50.50 & 66.10 & 70.90 & 78.70 & 78.20 & 90.80 & 85.30 & 84.20 & 61.80 & 63.80 & 68.20 & 69.80 & 67.20 & 73.70 \\
R$^3$Det-GWD~\cite{gwd} & 88.82 & 82.94 & 55.63 & 72.75 & 78.52 & 83.10 & 87.46 & 90.21 & 86.36 & 85.44 & 64.70 & 61.41 & 73.46 & 76.94 & 57.38 & 76.34 \\
R$^3$Det-KLD~\cite{kld} & 88.90 & 84.17 & 55.80 & 69.35 & 78.72 & 84.08 & 87.00 & 89.75 & 84.32 & 85.73 & 64.74 & 61.80 & 76.62 & 78.49 & 70.89 & 77.36 \\
S$^2$A-Net~\cite{s2anet} & 89.11 & 82.84 & 48.37 & 71.11 & 78.11 & 78.39 & 87.25 & 90.83 & 84.90 & 85.64 & 60.36 & 62.60 & 65.26 & 69.13 & 57.94 & 74.12 \\
RoI Transformer~\cite{roi_transformer} & 89.01 & 77.48 & 51.64 & 72.07 & 74.43 & 77.55 & 87.76 & 90.81 & 79.71 & 85.27 & 58.36 & 64.11 & 76.50 & 71.99 & 54.06 & 74.05 \\
ReDet~\cite{redet} & 88.79 & 82.64 & 53.97 & 74.00 & 78.13 & 84.06 & 88.04 & 90.89 & \textbf{87.78} & 85.75 & 61.76 & 60.39 & 75.96 & 68.07 & 63.59 & 76.25 \\
O-RCNN~\cite{orcnn} & 89.46 & 82.12 & 54.78 & 70.86 & 78.93 & 83.00 & 88.20 & \textbf{90.90} & 87.50 & 84.68 & 63.97 & 67.69 & 74.94 & 68.84 & 52.28 & 75.87 \\
ARC~\cite{arc} & 89.40 & 82.48 & 55.33 & 73.88 & 79.37 & 84.05 & 88.06 & \textbf{90.90} & 86.44 & 84.83 & 63.63 & \textbf{70.32} & 74.29 & 71.91 & 65.43 & 77.35 \\
GRA~\cite{gra} & -- & -- & -- & -- & -- & -- & -- & -- & -- & -- & -- & -- & -- & -- & -- & 77.63 \\
LSKNet~\cite{lsknet} & 89.66 & 85.52 & 57.72 & 75.70 & 74.95 & 78.69 & 88.24 & 90.88 & 86.79 & 86.38 & 66.92 & 63.77 & 77.77 & 74.47 & 64.82 & 77.49 \\
MutDet~\cite{mutdet} & 87.30 & 78.70 & 51.30 & 68.50 & 78.90 & 81.60 & 88.10 & 90.70 & 79.90 & 83.70 & 58.00 & 61.80 & 76.50 & 72.10 & 60.80 & 74.50 \\
PKINet~\cite{pkinet} & 89.72 & 84.20 & 55.81 & 77.63 & 80.25 & 84.45 & 88.12 & 90.88 & 87.57 & 86.07 & 66.86 & 70.23 & 77.47 & 73.62 & 62.94 & 78.39 \\
PKINet-v2-S~\cite{pkinetv2} & 89.79 & 83.87 & 55.18 & 73.52 & 74.88 & 84.26 & 88.47 & 90.87 & 87.45 & 86.49 & 66.22 & 62.00 & 78.32 & 69.55 & 68.01 & 77.26 \\
ReDiffDet (R50)~\cite{rediffdet} & 85.45 & 77.84 & 50.39 & 71.63 & 80.90 & 84.85 & 88.77 & 90.88 & 87.00 & 86.08 & 63.30 & 63.43 & 76.01 & 76.03 & 60.16 & 76.18 \\
ReDiffDet (Swin-T)~\cite{rediffdet} & 82.89 & 82.90 & 52.47 & 70.80 & 80.48 & 85.41 & \textbf{88.97} & \textbf{90.90} & 86.97 & \textbf{87.35} & 60.24 & 65.01 & 76.61 & 74.28 & 64.07 & 76.62 \\
Strip R-CNN~\cite{striprcnn} & 89.68 & 85.62 & 56.32 & 75.42 & 75.85 & 85.40 & 88.59 & \textbf{90.90} & 87.25 & 87.09 & \textbf{69.33} & 62.49 & 78.23 & 74.87 & \textbf{73.33} & 78.69 \\
O-RCNN + FAA~\cite{faa} & 89.40 & 84.15 & 54.66 & 72.23 & 78.42 & 82.89 & 87.86 & 90.89 & 86.26 & 85.24 & 63.95 & 65.58 & 75.30 & 70.44 & 61.06 & 76.55 \\
LSKNet + FAA~\cite{faa} & 89.85 & 85.07 & 55.39 & \textbf{77.77} & 79.30 & 84.72 & 88.56 & \textbf{90.90} & 87.74 & 86.29 & 68.37 & 67.74 & 77.19 & 74.14 & 64.30 & 78.49 \\
Strip R-CNN + FAA~\cite{faa} & 89.81 & 83.27 & 52.93 & 77.06 & 79.57 & 84.90 & 88.31 & 90.89 & 87.10 & 86.43 & 66.43 & 68.04 & 75.54 & 79.03 & 71.48 & 78.72 \\
\midrule
O-RCNN + NIF & 89.24 & 83.75 & 56.52 & 74.15 & 75.36 & 84.69 & 88.31 & 90.80 & 86.73 & 85.78 & 68.45 & 66.10 & 78.35 & 79.88 & 61.62 & 77.98 \\
PKINet-v2-S + NIF & \textbf{89.87} & 85.54 & \textbf{59.07} & 76.93 & 73.80 & 84.60 & 88.38 & 90.89 & 87.46 & 86.48 & 65.65 & 61.97 & 78.59 & 70.31 & 73.21 & 78.18 \\
Strip R-CNN + NIF & 89.64 & 85.08 & 55.17 & 73.55 & 75.42 & \textbf{85.69} & 88.67 & \textbf{90.90} & 87.75 & 87.13 & 68.54 & 63.29 & \textbf{78.62} & \textbf{82.85} & 73.13 & 79.03 \\
LSKNet-S + NIF & 89.72 & \textbf{85.81} & 57.77 & 76.77 & \textbf{81.16} & 85.33 & 88.53 & \textbf{90.90} & 87.52 & 86.15 & 66.19 & 63.24 & 78.51 & 81.05 & 71.18 & \textbf{79.32} \\
\bottomrule
\end{tabular*}
\caption{\textbf{Comparisons with SOTA methods on the DOTA-v1.0 dataset} (AP in percent). PL, BD, BR, GTF, SV, LV, SH, TC, BC, ST, SBF, RA, HA, SP, and HC follow the standard DOTA category abbreviations. Results for PKINet-v2-S and Strip R-CNN are from our local reproductions.}
\label{tab:dota_per_class}
\end{table*}

\paragraph{DOTA-v1.0.}
Table~\ref{tab:dota_per_class} combines FAA comparison results~\cite{faa}
with our official evaluations. NIF reaches 77.98, 78.18, 79.03, and
79.32\% mAP with Oriented R-CNN, PKINet-v2-S, Strip R-CNN, and LSKNet-S,
respectively; LSKNet-S gives the strongest single-scale result in the table.
In matched local comparisons, PKINet-v2-S rises from 77.26 to 78.18\% and
Strip R-CNN from 78.69 to 79.03\%, gains of 0.92\% and 0.34\%.
For PKINet-v2-S, bridge, ground-track-field, and helicopter AP increase by
3.89\%, 3.41\%, and 5.20\%; Strip R-CNN gains 7.98\% AP on swimming-pool.
These categories span elongated structures, large
spatial extents, and compact objects, showing that the localization gains
are not confined to a single object geometry.

\paragraph{HRSC2016.}
Table~\ref{tab:hrsc} compares four locally reproduced carriers with
their NIF counterparts under matched settings. Following FAA~\cite{faa},
we report AP$_{50}$, AP$_{75}$, and COCO-style mAP under VOC 2007 and
VOC 2012 interpolation. NIF raises COCO-style mAP for all four carriers
by 2.82--4.96\% under VOC 2007 and 4.32--6.66\% under VOC
2012. AP$_{50}$ changes by at most 0.52\%, while AP$_{75}$ improves
by 0.56--1.19\% under VOC 2007 and 2.30--3.27\% under VOC 2012,
indicating that the gains concentrate at stricter
localization thresholds across both conventional and Strip heads. This
pattern is consistent with NIF refining box boundaries after the carrier has
already identified and coarsely localized an object.

Table~\ref{tab:hrsc} also reports full-test complexity. Because HRSC preserves
aspect ratio, its input areas vary across images; NIF adds 0.13M parameters,
0.76--1.25 GFLOPs, and 2.84--7.37 ms of latency across the four carriers.

\begin{table}[!t]
\centering
\small
\setlength{\tabcolsep}{2.2pt}
\resizebox{\columnwidth}{!}{%
\begin{tabular}{lrrrrrr}
\toprule
Method & Params (M) & GFLOPs & Latency (ms) & AP$_{50}$ & AP$_{75}$ & mAP \\
\midrule
\multicolumn{6}{c}{PASCAL VOC 2007} \\
\midrule
O-RCNN~\cite{orcnn} & 41.35 & 118.96 & 43.85 & 90.40 & 88.60 & 70.66 \\
O-RCNN + NIF & 41.48 & 119.85 & 50.43 & 90.51 & 89.70 & \textbf{74.33} \\
LSKNet-S~\cite{lsknet} & 30.96 & 101.54 & 53.47 & 90.50 & 88.20 & 69.80 \\
LSKNet-S + NIF & 31.09 & 102.30 & 56.33 & 90.33 & 89.39 & \textbf{73.93} \\
PKINet-v2-S~\cite{pkinetv2} & 30.71 & 101.54 & 65.04 & 90.30 & 88.40 & 69.53 \\
PKINet-v2-S + NIF & 30.84 & 102.35 & 67.88 & 90.30 & 89.25 & \textbf{74.49} \\
Strip R-CNN~\cite{striprcnn} & 45.12 & 194.51 & 74.49 & 90.50 & 88.90 & 71.01 \\
Strip R-CNN + NIF & 45.25 & 195.76 & 81.86 & 90.42 & 89.46 & \textbf{73.83} \\
\midrule
\multicolumn{6}{c}{PASCAL VOC 2012} \\
\midrule
O-RCNN~\cite{orcnn} & 41.35 & 118.96 & 43.85 & 97.80 & 89.20 & 72.64 \\
O-RCNN + NIF & 41.48 & 119.85 & 50.43 & 98.32 & 92.20 & \textbf{78.09} \\
LSKNet-S~\cite{lsknet} & 30.96 & 101.54 & 53.47 & 98.10 & 88.50 & 71.77 \\
LSKNet-S + NIF & 31.09 & 102.30 & 56.33 & 98.02 & 91.77 & \textbf{78.43} \\
PKINet-v2-S~\cite{pkinetv2} & 30.71 & 101.54 & 65.04 & 98.00 & 88.90 & 72.26 \\
PKINet-v2-S + NIF & 30.84 & 102.35 & 67.88 & 98.19 & 92.02 & \textbf{78.43} \\
Strip R-CNN~\cite{striprcnn} & 45.12 & 194.51 & 74.49 & 98.30 & 89.60 & 74.22 \\
Strip R-CNN + NIF & 45.25 & 195.76 & 81.86 & 98.20 & 91.90 & \textbf{78.54} \\
\bottomrule
\end{tabular}
}
\caption{\textbf{Comparisons with SOTA methods on HRSC2016} (AP in \%). Models use trainval/test. Latency is synchronized end-to-end inference, including rotated RoIAlign, NIF, decoding, and NMS, on one RTX 2080 Ti.}
\label{tab:hrsc}
\end{table}

\begin{table}[H]
\centering
\scriptsize
\setlength{\tabcolsep}{2.8pt}
\begin{tabular}{@{}c l r r r@{}}
\toprule
Grp. & Variant & AP$_{50}$ & AP$_{90}$ & mAP \\
\midrule
A & Detector only & 90.50 & 23.70 & 71.01 \\
B & Direct 5D + box loss & 90.50 & 41.40 & 73.93 \\
C & Direct 5D + field loss & 90.50 & 41.60 & 73.96 \\
D & 20 obs. + uniform fit & 90.50 & 42.10 & 74.04 \\
E & D + global pooling & 90.50 & 35.20 & 73.15 \\
F & D $-$ observation loss & 90.50 & 42.00 & 74.01 \\
G & D $-$ box loss & 90.50 & 36.50 & 73.36 \\
H & NIF + reliability weighting & 90.50 & 42.40 & 74.15 \\
\bottomrule
\end{tabular}
\caption{\textbf{Controlled ablations on HRSC2016.} All rows use the same frozen Strip R-CNN carrier and matched trainval/test protocol. AP uses VOC 2007; mAP is the corresponding COCO-style mean. A is the detector-only baseline; B--H use the same local strip features and differ only in the listed refinement design or loss.}
\label{tab:hrsc_ablation}
\end{table}

\paragraph{Controlled ablations.}
Table~\ref{tab:hrsc_ablation} uses one frozen Strip R-CNN carrier, identical
strip features, and a matched budget. A is the detector-only baseline. B
keeps the features but adds a direct five-variable head, giving 41.40\%
AP$_{90}$ and 73.93\% mAP; this measures the benefit of extra regression
capacity alone. Adding field supervision further raises both metrics to
41.60\% and 73.96\% (C), showing that local boundary targets complement
direct box regression. Replacing the direct head with 20 observations and
uniform analytical fitting reaches 42.10\% AP$_{90}$ and 74.04\% mAP (D),
supporting the value of distributed measurements and explicit rectangle
compatibility beyond a learned five-coordinate update.

The structural choices are necessary under strict overlap. Global pooling
removes the ordered normal-position profile and lowers the result to 35.20\%
AP$_{90}$ and 73.15\% mAP (E), demonstrating the value of retaining
position-specific boundary evidence. Removing observation loss lowers both
scores relative to D (F: 42.00\% and 74.01\%), confirming that direct
supervision of the predicted field contributes to the final localization.
Removing the refinement-box loss produces a larger decline (G: 36.50\% and
73.36\%), establishing the role of supervision on the decoded rectangle.
Reliability learning and weighted fitting yield the best scores (H: 42.40\%
and 74.15\%), providing a further positive contribution over uniform
fitting. Together, the controlled comparisons support a positive role for
each component in the tested configuration. AP$_{50}$ remains 90.50\% in all
rows, locating these effects in box localization rather than classification.

\paragraph{FAIR1M-v1.0.}
Table~\ref{tab:fair1m} reports 46.36, 48.13, and 48.22\% mAP for
O-RCNN, LSKNet-S, and Strip R-CNN with NIF. The matched Strip R-CNN
improves from 47.92 to 48.22\%, a 0.30\% gain; O-RCNN rises from 45.60
to 46.36\% and LSKNet-S from 47.87 to 48.13\%, giving gains of 0.76\%
and 0.26\%. Scores use the 34 fine-grained categories. Because NIF only
refines proposals and leaves classification and regression paths unchanged,
the gains persist under fine-grained labels and multi-scale inference.

\begin{table*}[t]
\centering
\scriptsize
\setlength{\tabcolsep}{2.6pt}
\begin{tabular*}{\textwidth}{@{\extracolsep{\fill}}lccccccccccc}
\toprule
Method & \shortstack{G. V.\textsuperscript{*}\\\cite{gliding}} & \shortstack{RetinaNet\textsuperscript{*}\\\cite{retinanet}} & \shortstack{C-RCNN\textsuperscript{*}\\\cite{cascade_rcnn}} & \shortstack{F-RCNN\textsuperscript{*}\\\cite{fasterrcnn}} & \shortstack{RoI Trans.\textsuperscript{*}\\\cite{roi_transformer}} & \shortstack{O-RCNN\\\cite{orcnn}} & \shortstack{LSKNet-S\\\cite{lsknet}} & \shortstack{Strip R-CNN\\\cite{striprcnn}} & \shortstack{O-RCNN\\+ NIF} & \shortstack{LSKNet-S\\+ NIF} & \shortstack{Strip R-CNN\\+ NIF} \\
\midrule
mAP (\%) & 29.92 & 30.67 & 31.18 & 32.12 & 35.29 & 45.60 & 47.87 & 47.92 & \textbf{46.36} & \textbf{48.13} & \textbf{48.22} \\
\bottomrule
\end{tabular*}
\caption{\textbf{Comparisons with SOTA methods on the FAIR1M-v1.0 dataset}. *: Results are referenced from the FAIR1M paper~\cite{fair1m}.}
\label{tab:fair1m}
\end{table*}

\paragraph{DIOR-R.}
Table~\ref{tab:dior} reports 66.05\%, 68.19\%, and 67.46\% mAP with
O-RCNN, LSKNet-S, and Strip R-CNN plus NIF. The matched Strip R-CNN
gain is 0.71\% over its 66.75\% local baseline. Relative to the tabulated
O-RCNN and LSKNet-S results, the NIF entries are higher by 1.75\% and
2.29\%, respectively; gains across all three carriers hold under the native
DIOR-R test protocol.

The local rows use all 11,738 test images. NIF adds 0.13M parameters,
1.77--1.91 GFLOPs and 6.32--9.97 ms per image over the measured carrier timings;
the literature-only rows have no local latency measurement.

\begin{table}[t]
\centering
\small
\setlength{\tabcolsep}{2.8pt}
\resizebox{\columnwidth}{!}{%
\begin{tabular}{lrrrr}
\toprule
Method & Params (M) & GFLOPs & Latency (ms) & mAP (\%) \\
\midrule
RetinaNet-O~\cite{retinanet} & -- & -- & -- & 57.55 \\
Faster R-CNN-O~\cite{fasterrcnn} & 41.10 & 198.00 & -- & 59.54 \\
TIOE-Det~\cite{tioe} & 41.10 & 198.00 & -- & 61.98 \\
O-RCNN~\cite{orcnn} & 41.37 & 148.40 & 49.13 & 64.30 \\
ARS-DETR~\cite{arsdetr} & 41.10 & 198.00 & -- & 66.12 \\
O-RepPoints~\cite{reppoints} & 36.60 & -- & -- & 66.71 \\
DCFL~\cite{dcfl} & -- & -- & -- & 66.80 \\
LSKNet-S~\cite{lsknet} & 30.98 & 125.36 & 57.10 & 65.90 \\
PKINet-S~\cite{pkinet} & 30.86 & 132.20 & 227.43 & 67.03 \\
Strip R-CNN~\cite{striprcnn} & 45.14 & 218.05 & 80.68 & 66.75 \\
O-RCNN + NIF & 41.50 & 150.24 & 59.10 & \textbf{66.05} \\
LSKNet-S + NIF & 31.11 & 127.13 & 63.42 & \textbf{68.19} \\
Strip R-CNN + NIF & 45.27 & 219.96 & 90.31 & \textbf{67.46} \\
\bottomrule
\end{tabular}
}
\caption{\textbf{Comparisons with SOTA methods on DIOR-R}. AP is in percent. Latency is synchronized end-to-end inference, including rotated RoIAlign, NIF, decoding, and NMS, on one RTX 2080 Ti.}
\label{tab:dior}
\end{table}

\paragraph{Cross-framework transfer.}
We attach NIF to four DOTA-v1.0 carriers: Ultralytics YOLO26
OBB~\cite{yolo26}, O2-RTDETR~\cite{o2rtdetr}, LSKNet-S + FAA, and PKINet-v2-S.
Each starts from an official single-scale checkpoint; all detector
parameters are frozen and only NIF is optimized. YOLO26-L OBB uses
Ultralytics' official settings, while the others follow Table~\ref{tab:dota_per_class}.
Table~\ref{tab:cross_framework} shows AP$_{75}$ gains of 0.92\%, 1.88\%, 2.16\%,
and 1.62\% and mAP gains of 0.20\%, 1.14\%, 1.82\%, and 1.21\%,
respectively, showing that NIF reuses coarse boxes without detector changes
or retraining and delivers gains in the high-AP regime.

\begin{table}[!b]
\centering
\small
\setlength{\tabcolsep}{2.8pt}
\begin{tabular}{lrrrr}
\toprule
 & \multicolumn{2}{c}{Frozen carrier} & \multicolumn{2}{c}{Carrier + NIF} \\
Carrier & AP$_{75}$ & mAP & AP$_{75}$ & mAP \\
\midrule
YOLO26-L OBB~\cite{yolo26} & 59.19 & 54.18 & \textbf{60.11} & \textbf{54.38} \\
O2-RTDETR~\cite{o2rtdetr} & 54.28 & 51.09 & \textbf{56.16} & \textbf{52.23} \\
LSKNet-S + FAA~\cite{lsknet,faa} & 51.88 & 48.35 & \textbf{54.04} & \textbf{50.17} \\
PKINet-v2-S~\cite{pkinetv2} & 54.58 & 50.83 & \textbf{56.20} & \textbf{52.04} \\
\bottomrule
\end{tabular}
\caption{\textbf{Cross-framework transfer on DOTA-v1.0}. Each carrier is initialized from its official single-scale checkpoint, with all detector parameters frozen and only NIF optimized. AP$_{75}$ is evaluated at IoU $0.75$; mAP is the COCO-style mean AP averaged over IoU thresholds from $0.50$ to $0.95$. ``LSKNet-S + FAA'' denotes the FAA Head carrier.}
\label{tab:cross_framework}
\end{table}

\subsection{Does the Field Correct Geometry?}
\label{sec:mechanism}

\begin{figure}[t]
  \centering
  \includegraphics[width=\columnwidth]{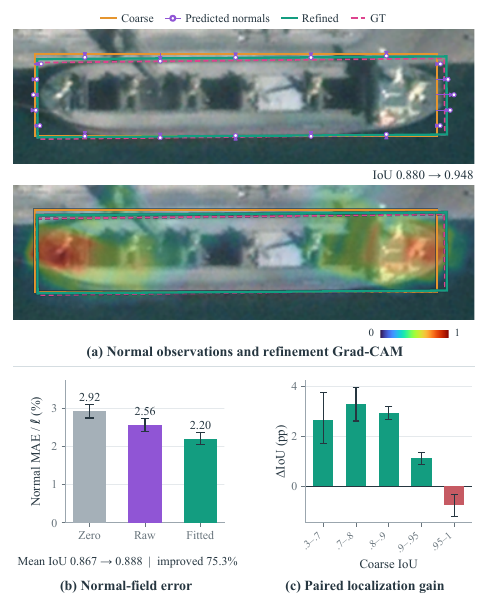}
  \caption{\textbf{Geometric refinement evidence on HRSC2016.}
  (a) Normal observations, box refinement, and Grad-CAM for the final update.
  (b) Normal-field error. (c) Paired IoU gain by coarse IoU.}
  \label{fig:mechanism}
\end{figure}

\paragraph{A fixed candidate cohort.}
We evaluate Oriented R-CNN NIF before NMS on all 444 HRSC test images to
isolate geometry from ranking. Per GT, we select its highest-scoring match
(score $\geq0.5$, coarse IoU $\geq0.3$; ties by index), regardless of refined
IoU or gain. Of 1,228 GTs, 1,212 have a match; none are synthetically added.

Figure~\hyperref[fig:mechanism]{\ref*{fig:mechanism}(a)} shows the 20 normal
observations, refinement, and output-conditioned Grad-CAM for one candidate;
\hyperref[fig:mechanism]{\ref*{fig:mechanism}(b)} compares zero, raw, and
projected fields; and \hyperref[fig:mechanism]{\ref*{fig:mechanism}(c)} groups
paired IoU gains by coarse IoU. Error bars use 2,000 image-cluster bootstrap
resamples.

Fig.~\hyperref[fig:mechanism]{\ref*{fig:mechanism}(a)} shows a joint side
correction that raises example IoU from 0.880 to 0.948, with Grad-CAM
focused near mismatched ship boundaries. In (b), normalized field MAE falls
from 0.02917 (zero) to 0.02565 (raw) and 0.02203 (projected), a 24.5\%
reduction, showing that projection improves consistency before decoding. In
(c), mean candidate IoU rises from 0.8669 to 0.8884 (2.148 points, 95\% CI
$[1.968,2.322]$), and 75.3\% of pairs improve; gains are positive in most
bins. The error is measured against the supporting-line field in
Eq.~\ref{eq:intersection}, rather than image segmentation edges.

\section{Conclusion}
\enlargethispage{4\baselineskip}

We present NIF, a plug-and-play normal-intersection field
for oriented-box refinement. It samples coarse-box normal
strips, predicts signed boundary intersections and reliabilities, and
analytically fits a compatible five-degree-of-freedom rectangle.
Tangential pooling preserves normal-position structure, while shared
projection limits the added parameters to 0.132M and leaves each carrier
unchanged. Across multiple public datasets,
NIF consistently improves high-IoU localization across detectors with
different architectures. Frozen-detector and cross-framework experiments
show that existing coarse boxes can be reused without detector
retraining; the ablations, field-error measurements, and candidate-IoU
analysis support distributed observations, compatibility fitting, and
detector-independent transfer.

\section*{Acknowledgments}
We thank Shanghai Zhongchao Automation Engineering Co., Ltd. for providing
the computational resources that supported this work.

{\small
\bibliographystyle{ieeenat_fullname}
\bibliography{references}
}

\clearpage
\section*{Supplementary Material}
\section{NIF Implementation}
\subsection{Geometry and local observations}
For a coarse oriented box $P=(\mathbf c,2a,2b,\theta)$, $\mathbf c$ is its center, $a,b$ are its half-sizes, and $\theta$ is its orientation. NIF places five equally spaced anchors along each supporting side at normalized tangential positions $s_j\in[-0.75,0.75]$, giving $4\times5=20$ observations per RoI. At each anchor, a $3\times9$ strip is sampled from an FPN feature map. The three tangent offsets are $-d,0,d$, where $d$ is the assigned FPN stride in image pixels. Nine equally spaced normal offsets span $[-R_i,R_i]$, with $R_i=\max(a/2,2d)$ for sides normal to the width axis and $R_i=\max(b/2,2d)$ for sides normal to the height axis. Thus $0.75$ controls anchor placement along a side, while the local strip extent depends on the box half-size and FPN stride. Sampling uses the same aligned image-to-feature coordinate convention as the RoI extractor and zero padding outside the feature map.

The FPN levels share a $1\times1$ projection from 256 to 64 channels. Two $3\times3$ convolutions operate on the projected strip features. Features are averaged only along the tangent direction, preserving the ordered nine normal positions. A small multilayer perceptron combines this visual descriptor with seven normalized geometric quantities: edge type, side sign, tangential position, the two normalized half-sizes, the normalized strip radius, and the normalized sampling stride. It predicts one normal displacement and, in the reliability-aware setting, one reliability value for each observation.

The predicted observations are converted to a five-degree-of-freedom box by a differentiable supporting-line least-squares fit. Following the main paper, $J(P)\in\mathbb R^{20\times5}$ is the design matrix induced by the four sides, $\hat{\mathbf y}$ contains the normalized predicted normal offsets, and $\mathbf z\in\mathbb R^5$ contains the normalized translation, size, and rotation variables. We fit
\begin{equation}
    \hat{\mathbf z}=\arg\min_{\mathbf z}
    \left\|\mathbf W^{1/2}\bigl(J(P)\mathbf z-\hat{\mathbf y}\bigr)\right\|_2^2.
\end{equation}
Here, $\mathbf W$ is diagonal with weights proportional to $r_i^{1.5}$ and normalized to mean one within each RoI; $r_i\in(0.05,1)$ is the predicted reliability. Gradients pass through both the observations and reliabilities in this fit. The fitted variables are analytically decoded to the refined box $P^+$. The linear system is solved in FP32, including when mixed-precision training is enabled.

\subsection{Losses and reliability targets}
For each sampled positive RoI, the target $y_i^*=h_i^*/\ell$ is the signed normal displacement $h_i^*$ from its anchor to the matched ground-truth supporting line, normalized by the coarse-box half-diagonal $\ell=\sqrt{a^2+b^2}$. We add
\begin{equation}
 \mathcal{L}_{\mathrm{NIF}}=\mathcal{L}_{\mathrm{obs}}+\mathcal{L}_{\mathrm{ref}}+0.05\,\mathcal{L}_{\mathrm{rel}},
\end{equation}
with
\begin{align}
 \mathcal L_{\mathrm{obs}}
 &=\mathbb E_P\!\left[\frac1{20}\sum_{i=1}^{20}
 \bar w_i(\hat y_i-y_i^*)^2\right],\\
 \bar w_i&=\frac{\operatorname{sg}(r_i^{1.5})}
 {\frac1{20}\sum_{i'=1}^{20}\operatorname{sg}(r_{i'}^{1.5})},\\
 \mathcal L_{\mathrm{ref}}
 &=\mathbb E_{P,G}[1-\operatorname{rIoU}(P^+,G)].
\end{align}
Here, $\hat y_i$ is a predicted normalized displacement, $i$ and $i'$ index the 20 observations, and the expectations are empirical means over selected RoIs and their matched ground-truth boxes $G$. The operator $\operatorname{sg}$ preserves its forward value and blocks its derivative. Observation-loss weights are detached to prevent this loss from directly lowering reliability, and normalized to mean one separately within each RoI. The refinement term uses the intersection area divided by the union area of $P^+$ and $G$, computed from exact polygon intersections with piecewise differentiable geometry.

The reliability target is computed from a detached observation error,
\begin{equation}
 r_i^*=\operatorname{clip}\!\left(\exp\!\left[-\frac{|\operatorname{sg}(\hat y_i)-y_i^*|}{0.25}\right],0.05,1\right),
\end{equation}
where clipping bounds the target to $[0.05,1]$. We optimize $\mathcal L_{\mathrm{rel}}$ as the mean smooth-$L_1$ loss between $r_i$ and $r_i^*$ over RoIs and observations, with transition parameter $\beta_r=0.1$. The stop-gradient prevents the reliability target from changing the displacement predictor through this auxiliary term. The reliability prior is $0.8$, so the freshly initialized branch starts with a useful nonzero weight for every observation.

\subsection{Proposal sampling}
NIF reads the detector's sampled positive proposals and leaves the detector's original classification and regression path unchanged. At most 128 proposals per image are used for NIF supervision. For the reported low-IoU-aware protocol, proposals are assigned to the intervals $[0.30,0.45)$, $[0.45,0.60)$, $[0.60,0.75)$, $[0.75,0.90)$, and $[0.90,1.01)$ before the budget is applied. Ground-truth-injected proposals are excluded whenever ordinary positives are available; a ground-truth fallback keeps the branch connected on images without eligible ordinary positives. Inference applies NIF after coarse box regression and before the detector's final rotated NMS. The NIF branch is therefore a plug-in refinement step: it adds parameters and local computation but does not replace the carrier's prediction head.

\subsection{Decoding validity and clamp frequency}
The five-dimensional projection enforces the linear relationships among supporting-line observations. Positivity of the decoded dimensions requires additional inequalities. Write the fitted variables as $\hat{\mathbf z}=(z_x,z_y,z_a,z_b,z_t)^\top$ and let $\hat t=z_t/\kappa$, where $\kappa=(\frac15\sum_{j=1}^5 s_j^2)^{1/2}$. Before clamping, the decoded full sizes are
\begin{equation}
 \check w=\frac{2(a+\ell z_a)}{\sqrt{1+\hat t^2}},
 \qquad
 \check h=\frac{2(b+\ell z_b)}{\sqrt{1+\hat t^2}}.
\end{equation}
Here, $\check w,\check h$ denote the raw width and height in pixels. Since the denominator is positive, a nondegenerate rectangle requires $a+\ell z_a>0$ and $b+\ell z_b>0$; unconstrained least squares does not enforce these inequalities. The implementation outputs $w^+=\max(\check w,\epsilon)$ and $h^+=\max(\check h,\epsilon)$, with $\epsilon=10^{-3}$ pixels. A candidate counts as a clamp event if either raw dimension is below $\epsilon$, before long-edge canonicalization.

Exact recovery applies when the observations equal the target field on the consistently selected target-axis branch, the design matrix has full column rank, and all fitting weights are positive. The target dimensions must be at least $\epsilon$ so that the size safeguard leaves them unchanged. Under these conditions, in exact arithmetic, the weighted fit recovers the target five-vector and analytic decoding recovers its rectangle, up to equivalent box parameterizations. With noisy predicted observations, the fitted field retains its linear structure, but exact target recovery and an IoU improvement for every candidate are not guaranteed. If clamping changes a decoded size, the output rectangle's supporting-line field can differ from the pre-clamp projected field.

We audit the Oriented R-CNN+NIF epoch-36 checkpoint used for the main paper's HRSC2016 normal-field analysis. Replaying the FP32 decoder on all 444 cached test images gives zero clamp events among 42,323 native candidates entering NIF before final rotated NMS (0\%). The 1,212 matched candidates in the analysis likewise contain zero clamp events (0\%). The smallest raw full dimension across all candidates is 9.19 pixels. The replay exactly reproduces the cached refined boxes. These frequencies refer to this checkpoint and test split.

\subsection{DIOR-R Reliability Diagnostic}
We directly evaluate whether the reliability branch ranks observation quality on
DIOR-R. We use the LSKNet-S+NIF epoch-12 checkpoint and the native test split;
the analysis contains 2,734,012 NIF candidates from 11,652 images and
54,680,240 observations. For each observation, we compare its predicted
reliability $r_i$ with the absolute normalized field error
$e_i=|\hat y_i-y_i^*|$. The Spearman correlation between $r_i$ and $e_i$ is
$-0.5275$ (equivalently, $+0.5275$ with the error-derived reliability target),
and the overall mean absolute field error is $0.0900$. Table~\ref{tab:dior_reliability}
groups observations into equal-width reliability intervals. The mean field
error falls from $0.3302$ in the lowest-weight interval to $0.0298$ in the
highest-weight interval. Thus, the reliability branch does rank observation
quality: high-weight observations usually have lower field error, providing a
direct basis for the error-aware weighted analytic fit.

\begin{table}[t]
\centering
\scriptsize
\setlength{\tabcolsep}{3pt}
\caption{DIOR-R reliability diagnostic. Each row is an equal-width interval
of predicted reliability; $N$ is the number of observations and MAE is the
mean absolute normalized field error.}
\label{tab:dior_reliability}
\begin{tabular}{c r r r}
\toprule
Reliability range & $N$ & $\bar r$ & MAE \\
\midrule
{[.050,.145)} & 896 & .128 & .3302 \\
{[.145,.240)} & 17,154 & .209 & .3319 \\
{[.240,.335)} & 114,590 & .300 & .3166 \\
{[.335,.430)} & 504,840 & .393 & .2781 \\
{[.430,.525)} & 1,576,707 & .485 & .2420 \\
{[.525,.620)} & 3,615,335 & .578 & .2080 \\
{[.620,.715)} & 5,922,365 & .670 & .1648 \\
{[.715,.810)} & 9,516,954 & .767 & .1091 \\
{[.810,.905)} & 19,545,021 & .864 & .0603 \\
{[.905,1.000]} & 13,866,378 & .928 & .0298 \\
\bottomrule
\end{tabular}
\end{table}

\subsection{Reliability intervention}
We use the same 1,212 matched HRSC2016 candidates as the main paper's normal-field analysis and change only the inference-time fitting weights, keeping the candidate boxes and network predictions fixed. Replacing predicted fitting weights by uniform weights gives a 2.134-point mean IoU gain, versus 2.148 points with learned weights (difference 0.014, 95\% interval $[0.009,0.020]$), with all parameters fixed. The confidence interval uses 2,000 image-cluster bootstrap resamples.

\section{Inference-Time Technical Details}
At inference, class-wise rotated NMS at IoU 0.6 selects representative coarse boxes for NIF refinement. Unselected candidates retain their coarse boxes and original scores, and all candidates continue through the carrier's regular score filtering and final rotated NMS. For DIOR-R, the shared settings are a score threshold of 0.05, final rotated-NMS IoU of 0.1, RPN-NMS IoU of 0.8, and limits of 2,000 RPN proposals and 2,000 final detections per image.

The inference path constructs strip coordinates for multiple RoIs in a batched operation, projects each active FPN feature map once per image and level, and processes RoIs in chunks of 256. Reliability-weighted fitting uses an algebraically equivalent weighted-moment form: it eliminates the four edge intercepts and solves for the shared slope through the Schur complement, avoiding explicit construction of the design matrix and a batched $5\times5$ linear solve.

HRSC2016 timing and complexity cover all 444 test images using the aspect-preserving 800-pixel test pipeline; DIOR-R covers all 11,738 native $800\times800$ test images. Both use FP32, batch size one, and one RTX 2080 Ti per model. Timing is synchronized, excludes ten warm-up images, and includes the complete detector forward, rotated RoIAlign, NIF sampling and fitting, decoding, NMS, and output transfer; data loading, evaluation, and file writing are excluded. GFLOPs are counted in a separate full-test pass using supported MMCV modules, with one multiply-add counted as one FLOP. Functional strip sampling, geometric fitting, rotated RoIAlign, and NMS are all included in latency.

\section{MMRotate Training Details}
All two-stage experiments use the MMRotate implementation and AdamW. Unless stated otherwise in the main paper, the learning rate is $10^{-4}$, weight decay is $0.05$, and AdamW uses $(\beta_1,\beta_2)=(0.9,0.999)$. Four-GPU runs use two images per GPU; the learning rate is kept at $10^{-4}$ rather than linearly scaled with the number of GPUs. The random seed is 0. Checkpoints are written every epoch and the final test protocol uses the specified final checkpoint or the dataset-specific selection rule described in the main paper.

For DOTA-v1.0, trainval images are split into $1024\times1024$ patches with 200-pixel overlap and the official test images are processed with the same patching convention. The 36-epoch schedule uses warm-up for 500 iterations (warm-up ratio $1/3$) and learning-rate drops at epochs 24 and 33. HRSC2016 uses the standard 617-image trainval split and 444-image test split; the longer image side is resized to 800 pixels while preserving aspect ratio, with the same 36-epoch schedule and drops at epochs 24 and 33. These details describe the protocol only; no additional DOTA or HRSC result table is introduced here.

DIOR-R uses $800\times800$ inputs, 12 epochs, learning-rate drops at epochs 8 and 11, horizontal/vertical/diagonal flips with ratio $0.25$, and PolyRandomRotate with ratio $0.5$ and angle range $180^\circ$. The native DIORDataset evaluation uses rotated IoU $0.5$ and the VOC2007 eleven-point AP protocol.

FAIR1M-v1.0 contains 37 labels: 34 fine-grained scored classes and three ``other'' classes. We use the native multiscale preprocessing: image rates $\{0.5,1.0,1.5\}$, $1024\times1024$ patches, and 500-pixel overlap. Training lasts 12 epochs, with learning-rate reductions by a factor of ten at epochs 8 and 11, a 500-iteration linear warm-up with initial ratio $1/3$, and the native FAIR augmentation and class policy. The three ``other'' classes are retained by the detector but are not included in the reported fine-grained mAP.

\paragraph{Training regime for the main-paper tables.}
The local baseline and NIF rows in Tables 1, 2, 4, and 5 are fully trained end-to-end rather than frozen-detector experiments. In the NIF runs, gradients from the NIF losses propagate through the NIF branch, sampled RoI features, FPN, and backbone; the FPN and backbone are therefore not frozen. The carrier's own detection losses continue to update its original head, so all detector components and NIF are optimized jointly. The frozen-detector settings are used only for the explicitly identified ablation and cross-framework transfer experiments. For the local baseline/NIF pairs in Table 1, the two runs use exactly the same DOTA patch extraction and merge procedure, data augmentation, candidate/proposal budget, score filtering, rotated-NMS IoU threshold, and post-NMS candidate cap. Thus, the paired comparison changes the added NIF branch and its losses while keeping the detector and inference protocol fixed.

\section{Training Stability Across Random Seeds}
We evaluate run-to-run stability on DOTA-v1.0 using four independent random seeds (0--3) for LSKNet-S+NIF and Strip R-CNN+NIF. Each run uses the same trainval split, 36-epoch schedule, optimizer, augmentation, and checkpoint protocol, and is evaluated on the official test split. AP$_{50}$ and AP$_{75}$ denote AP at IoU thresholds 0.50 and 0.75, respectively; mAP is the COCO-style mean AP over IoU thresholds 0.50:0.05:0.95. Table~\ref{tab:seed_stability} reports every seed, the arithmetic mean, the unbiased sample variance (divisor $n-1$), and the sample standard deviation. Variances are in squared percentage points; scores and standard deviations are in percentage points.

\begin{table*}[t]
\centering
\small
\setlength{\tabcolsep}{8pt}
\caption{DOTA-v1.0 test-set results across four random seeds. Mean, sample variance, and sample standard deviation are computed independently for each metric.}
\label{tab:seed_stability}
\begin{tabular}{llrrr}
\toprule
Carrier & Seed / statistic & AP$_{50}$ & AP$_{75}$ & COCO-style mAP \\
\midrule
\multirow{7}{*}{LSKNet-S+NIF}
& Seed 0 & 79.32 & 56.00 & 51.18 \\
& Seed 1 & 79.23 & 55.87 & 51.17 \\
& Seed 2 & 79.29 & 55.95 & 51.13 \\
& Seed 3 & 79.23 & 55.76 & 51.13 \\
& Mean & 79.268 & 55.895 & 51.153 \\
& Sample variance & 0.002025 & 0.010967 & 0.000692 \\
& Sample std. dev. & 0.045 & 0.105 & 0.026 \\
\midrule
\multirow{7}{*}{Strip R-CNN+NIF}
& Seed 0 & 79.02 & 55.96 & 51.60 \\
& Seed 1 & 78.94 & 55.92 & 51.58 \\
& Seed 2 & 79.00 & 55.87 & 51.61 \\
& Seed 3 & 78.99 & 55.93 & 51.57 \\
& Mean & 78.988 & 55.920 & 51.590 \\
& Sample variance & 0.001158 & 0.001400 & 0.000333 \\
& Sample std. dev. & 0.034 & 0.037 & 0.018 \\
\bottomrule
\end{tabular}
\end{table*}

Across the four runs, the standard deviation is at most 0.105 percentage points for LSKNet-S+NIF and 0.037 for Strip R-CNN+NIF; the COCO-style mAP standard deviations are 0.026 and 0.018, respectively. The small spread across both carriers indicates that NIF training is stable with respect to these random-seed changes.

\section{Ultralytics YOLO26-OBB Training}
The end-to-end cross-framework experiment uses the Ultralytics YOLO26-OBB implementation. The carrier is initialized from the official \texttt{yolo26l-obb.pt} checkpoint and frozen with \texttt{freeze=1000}; only the NIF branch is optimized. The four-GPU run uses a global batch size of 32 (8 images per GPU), $1024$-pixel images, 50 epochs, AdamW, initial learning rate $0.001$, final learning-rate fraction $0.01$, weight decay $5\times10^{-4}$, three warm-up epochs, and seed 0. Validation uses the DOTA validation split every epoch and \texttt{max\_det=700}; the carrier's detector weights remain unchanged throughout training and validation.

The Ultralytics augmentation settings are mosaic $1.0$, mixup $0.15$, and mosaic closure during the last five epochs. AMP is enabled for the detector, while the NIF weighted least-squares solve is promoted to FP32. This separation avoids a half-precision solver failure without disabling mixed-precision feature extraction. The experiment is designed to test reuse of an existing detector: NIF receives its coarse OBBs and feature maps and learns only the local correction branch.

\section{DIOR-R Per-Class Results}
Table~\ref{tab:supp_dior} reports the native DIOR-R test measurements from the four available local runs. Values are AP at rotated IoU $0.5$ in percent; the final row is the arithmetic mean over the 20 categories. The table is intentionally per-class so that changes in the overall score can be traced to individual object types.

\begin{table*}[t]
\centering
\scriptsize
\setlength{\tabcolsep}{3.2pt}
\caption{DIOR-R per-class AP (\%). Native DIORDataset evaluation at rotated IoU $0.5$ with VOC2007 eleven-point AP.}
\label{tab:supp_dior}
\begin{tabular}{lrrrr}
\toprule
Class & Strip R-CNN & O-RCNN+NIF & LSKNet-S+NIF & Strip R-CNN+NIF\\
\midrule
Airplane & 71.90 & 71.58 & 80.84 & 71.88\\
Airport & 35.66 & 40.80 & 44.48 & 41.53\\
Baseball field & 80.23 & 72.00 & 80.24 & 79.95\\
Basketball court & 88.69 & 89.18 & 89.64 & 89.60\\
Bridge & 44.89 & 43.43 & 45.18 & 45.19\\
Chimney & 78.23 & 77.41 & 78.23 & 78.73\\
Dam & 34.46 & 31.59 & 34.39 & 33.91\\
Expressway-service area & 79.92 & 83.56 & 85.07 & 79.25\\
Expressway-toll station & 70.00 & 68.29 & 70.16 & 69.63\\
Golf field & 74.87 & 75.60 & 75.30 & 76.83\\
Ground track field & 84.13 & 83.11 & 83.71 & 83.85\\
Harbor & 44.59 & 43.13 & 44.84 & 44.08\\
Overpass & 57.96 & 58.21 & 57.40 & 58.78\\
Ship & 81.13 & 81.23 & 81.11 & 81.09\\
Stadium & 80.22 & 77.51 & 79.21 & 80.76\\
Storage tank & 70.64 & 70.42 & 69.86 & 70.93\\
Tennis court & 88.70 & 81.40 & 89.24 & 89.25\\
Train station & 53.43 & 58.42 & 58.61 & 58.28\\
Vehicle & 49.33 & 48.25 & 49.43 & 49.11\\
Windmill & 66.04 & 65.96 & 66.97 & 66.54\\
\midrule
mAP & 66.75 & 66.05 & 68.19 & 67.46\\
\bottomrule
\end{tabular}
\end{table*}

The LSKNet-S carrier gives the highest score among the listed DIOR-R runs, while the Strip R-CNN refinement changes the result from 66.75 to 67.46. Class-level variation is expected because DIOR-R contains both large structured regions (courts and fields) and small or cluttered categories (airports, dams, and harbors); NIF only changes the geometric localization stage and does not alter class scores.

\section{DIOR-R Refinement Qualitative Results}
\begin{figure*}[t]
\centering
\includegraphics[width=\textwidth]{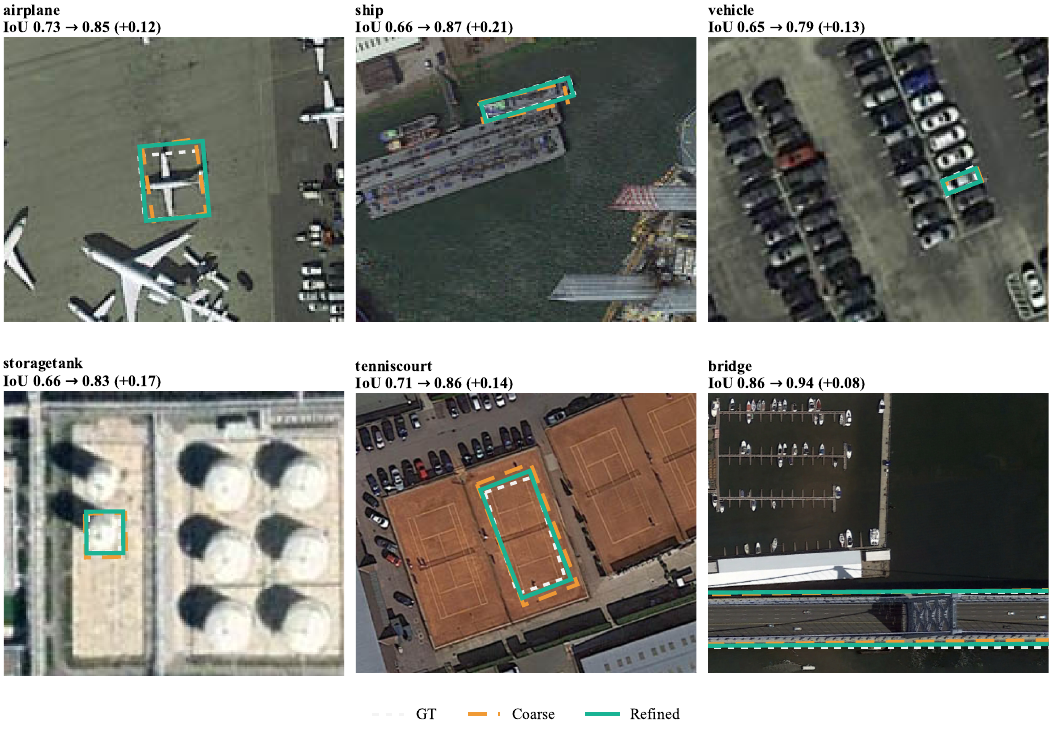}
\caption{Qualitative DIOR-R test-set refinements from LSKNet-S+NIF (epoch 12). White dashed, orange dashed, and green solid outlines represent the matched ground truth, the coarse detector box, and the NIF-refined box. Each panel reports the rotated IoU before and after refinement for the same candidate.}
\label{fig:dior_refinement_qualitative}
\end{figure*}

Figure~\ref{fig:dior_refinement_qualitative} shows six DIOR-R test examples processed by LSKNet-S+NIF at epoch 12. Each panel keeps one final-NMS-surviving candidate fixed while overlaying its matched ground-truth box, coarse detector box, and NIF-refined box. The white dashed, orange dashed, and green solid outlines denote GT, coarse, and refined boxes, respectively. The displayed IoU values increase from 0.73 to 0.85 (airplane), 0.66 to 0.87 (ship), 0.65 to 0.79 (vehicle), 0.66 to 0.83 (storage tank), 0.71 to 0.86 (tennis court), and 0.86 to 0.94 (bridge). The examples cover different object scales and orientations and provide a direct visual check of the localization correction measured by the per-class results above.

\section{FAIR1M-v1.0 Fine-Grained Results}
Table~\ref{tab:supp_fair} reproduces the complete fine-grained category table from the LSKNet FAIR1M benchmark. The first eight method columns are transcribed from that source. FAIR1M mAP is the mean of the 34 fine-grained categories.

\begin{table*}[t]
\centering
\scriptsize
\setlength{\tabcolsep}{1.6pt}
\renewcommand{\arraystretch}{0.91}
\caption{FAIR1M-v1.0 fine-grained AP (\%).}
\label{tab:supp_fair}
\resizebox{\textwidth}{!}{%
\begin{tabular}{llrrrrrrrrrrrr}
\toprule
Coarse & Fine class & GV$^*$ & Retina$^*$ & Cascade$^*$ & Faster$^*$ & ROI Trans.$^*$ & O-RCNN & LSKNet-T & LSKNet-S & Strip R-CNN & O-RCNN+NIF & LSKNet-S+NIF & Strip R-CNN+NIF\\
\midrule
\multirow{10}{*}{Airplane} & Boeing737 & 35.43&38.46&40.42&36.43&39.58&42.84&45.12&39.84&40.08&44.58&40.96&43.75\\
& Boeing747 & 47.88&55.36&52.86&50.68&73.56&87.61&84.97&86.63&85.82&85.94&85.58&86.29\\
& Boeing777 & 15.67&24.75&29.07&22.50&18.32&18.83&20.16&24.21&18.73&19.57&22.18&22.80\\
& Boeing787 & 48.32&51.81&52.47&51.86&56.43&62.92&56.00&56.48&54.97&60.36&54.70&61.51\\
& C919 & 0.01&0.81&0.00&0.01&0.00&22.17&25.77&24.17&23.12&20.99&28.07&27.32\\
& A220 & 40.11&40.50&44.37&47.81&47.67&47.87&50.05&52.20&49.04&48.59&51.16&50.62\\
& A321 & 39.31&41.06&38.35&43.83&49.91&70.25&71.63&73.31&73.00&71.08&74.08&72.86\\
& A330 & 16.54&18.02&26.55&17.66&27.64&73.34&67.94&72.82&72.57&69.20&72.07&71.95\\
& A350 & 16.56&19.94&17.54&19.95&31.79&77.19&74.04&75.83&77.43&75.12&78.48&76.34\\
& ARJ21 & 0.01&1.70&0.00&0.13&0.00&32.49&40.24&46.39&43.75&40.19&46.17&38.62\\
\midrule
\multirow{8}{*}{Ship} & Passenger Ship & 9.12&9.57&12.10&9.81&14.31&20.21&19.23&20.43&19.98&17.97&18.77&21.20\\
& Motorboat & 23.34&22.55&28.84&28.78&28.07&72.13&71.08&71.38&72.74&69.11&74.61&74.81\\
& Fishing Boat & 1.23&1.33&0.71&1.77&1.03&13.53&14.70&15.81&16.46&12.84&15.78&17.80\\
& Tugboat & 15.67&16.37&15.35&17.65&14.32&35.50&37.09&32.84&43.43&37.95&39.87&35.59\\
& Engineering Ship & 15.43&19.11&18.53&16.47&15.97&16.23&16.60&14.79&17.98&17.85&16.72&17.58\\
& Liquid Cargo Ship & 15.32&14.26&14.63&16.19&18.04&26.49&24.74&25.37&25.85&27.92&26.05&26.37\\
& Dry Cargo Ship & 25.43&24.70&25.15&27.06&26.02&38.43&40.57&41.29&39.20&39.60&39.93&40.05\\
& Warship & 13.56&15.37&14.53&13.16&12.97&34.74&38.70&36.20&39.76&32.54&40.97&40.48\\
\midrule
\multirow{9}{*}{Vehicle} & Small Car & 66.23&65.20&68.19&68.42&68.80&74.25&75.73&76.34&76.15&75.24&76.28&76.09\\
& Bus & 23.43&22.42&28.25&28.37&37.41&47.02&46.27&55.54&53.66&53.76&55.54&54.23\\
& Cargo Truck & 46.78&44.17&48.62&51.24&53.96&50.22&54.06&55.84&55.02&52.55&55.51&54.71\\
& Dump Truck & 36.56&35.37&40.40&43.60&45.68&57.56&59.52&61.57&60.28&58.80&61.05&60.27\\
& Van & 53.78&52.44&58.00&57.51&58.39&75.22&75.57&76.71&76.01&75.84&76.35&75.76\\
& Trailer & 14.32&19.17&13.66&15.03&16.22&20.91&19.30&21.46&21.30&21.05&21.38&22.76\\
& Tractor & 16.39&1.28&0.91&3.04&5.13&2.99&3.68&7.19&3.77&4.32&4.67&5.65\\
& Excavator & 16.92&17.03&16.45&17.99&22.17&19.95&28.40&25.73&26.57&28.25&27.04&29.93\\
& Truck Tractor & 28.91&28.98&30.27&29.36&46.71&1.77&5.66&4.74&8.31&5.87&3.61&7.16\\
\midrule
\multirow{4}{*}{Court} & Basketball Court & 48.41&50.58&38.81&58.26&54.84&55.35&59.74&61.78&60.13&57.22&60.08&59.61\\
& Tennis Court & 80.31&81.09&80.29&82.67&80.35&82.96&87.07&81.06&85.84&82.05&84.48&84.26\\
& Football Field & 53.46&52.50&48.21&54.50&56.68&64.62&69.67&70.39&71.41&69.23&68.98&68.64\\
& Baseball Field & 66.93&66.76&67.90&71.71&69.07&90.36&90.03&89.94&90.90&89.23&89.09&89.55\\
\midrule
\multirow{3}{*}{Road} & Intersection & 59.41&60.13&55.67&59.86&58.44&60.82&60.58&62.90&61.87&61.70&61.48&60.76\\
& Roundabout & 16.25&17.41&20.35&16.92&18.58&20.47&23.20&27.00&23.21&22.00&24.37&23.62\\
& Bridge & 10.39&12.58&12.62&11.87&31.81&33.40&38.57&39.51&40.91&27.67&40.47&40.55\\
\midrule
\multicolumn{2}{l}{mAP (34 classes)} & 29.92&30.67&31.18&32.12&35.29&45.60&46.93&47.87&47.92&46.36&48.13&48.22\\
\bottomrule
\end{tabular}
}%
\end{table*}

\section{FAIR1M-v1.0 Test-set Qualitative Results}
\begin{figure*}[t]
\centering
\includegraphics[width=\textwidth]{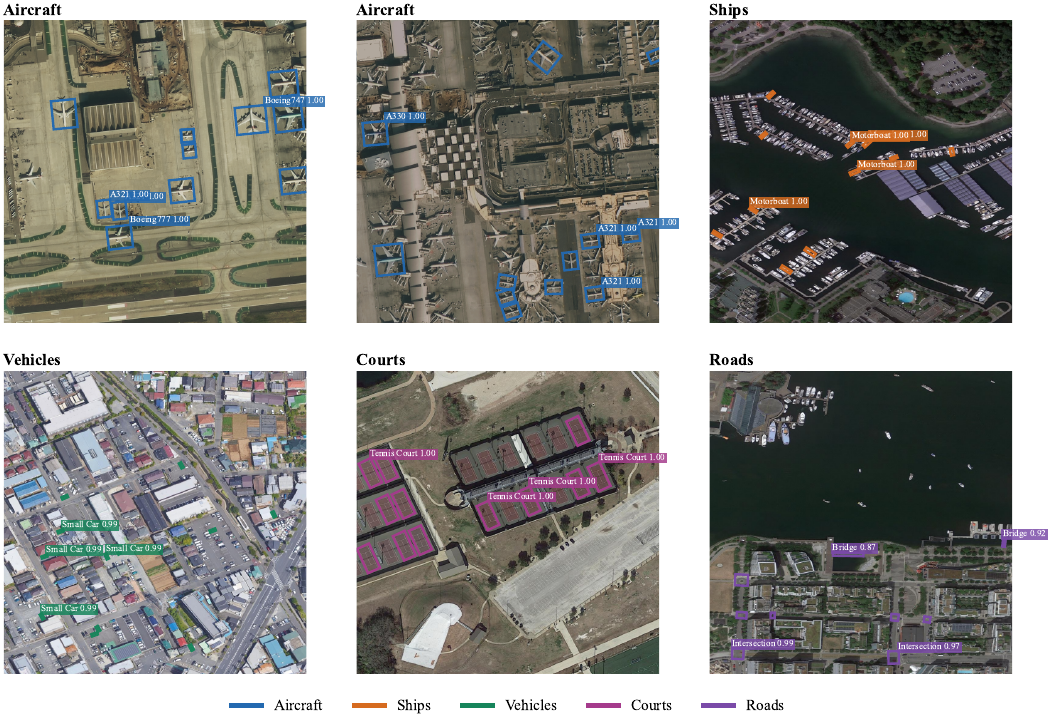}
\caption{Qualitative FAIR1M-v1.0 test-set predictions from LSKNet-S+NIF (epoch 12).}
\label{fig:fair_test_qualitative}
\end{figure*}

Figure~\ref{fig:fair_test_qualitative} shows six representative predictions from the FAIR1M-v1.0 test set. We use the LSKNet-S+NIF checkpoint from epoch 12 and the native multiscale test pipeline: each original image is tiled, predicted independently, and merged back to the original image coordinates. The examples cover aircraft, ships, vehicles, courts, and roads. These are test-set predictions without ground-truth overlays and are included as qualitative evidence of the detector's behavior under fine-grained labels and multiscale inference.

\end{document}